\documentclass[runningheads,a4paper]{llncs}
\usepackage{times}
\usepackage{amssymb}
\usepackage{graphicx}
\usepackage{url}
\usepackage{booktabs}
\usepackage{multirow}
\usepackage{subfig}
\usepackage{lscape}
\usepackage{makecell}
\usepackage{pifont}
\newcommand{\keywords}[1]{\par\addvspace\baselineskip
\noindent\keywordname\enspace\ignorespaces#1}

\begin{document}

\mainmatter  

\title{Lyric document embeddings for music tagging}

\titlerunning{Lyric Document Embeddings}

%
%
\author{Matt McVicar \and Bruno Di Giorgi \and Baris Dundar \and Matthias Mauch}
%
\authorrunning{Matt McVicar and Bruno Di Giorgi and Baris Dundar and Matthias Mauch}

\institute{Apple\\ \email{mmcvicar@apple.com}}

%
%

\maketitle

\begin{abstract}
We present an empirical study on embedding the lyrics of a song into a fixed-dimensional feature for the purpose of music tagging. Five methods of computing token-level and four methods of computing document-level representations are trained on an industrial-scale dataset of tens of millions of songs. We compare simple averaging of pretrained embeddings to modern recurrent and attention-based neural architectures. Evaluating on a wide range of tagging tasks such as genre classification, explicit content identification and era detection, we find that averaging word embeddings outperform more complex architectures in many downstream metrics.
\end{abstract}

\section{Introduction}\label{sec:intro}
Song lyrics have been shown to be effective predictors of emotion \cite{yang2009music}, and can be indicative of genre \cite{tsaptsinos2017lyricsbased,fell2014lyrics,ying2012genre,mayer2011musical,mckay2010evaluating,mayer2008rhyme,mayer2008combination}, mood \cite{hu2010lyrics,ying2012genre,hu2010improving,hu2009lyric,laurier2008multimodal,delbouys2018music}, music exploration \cite{watanabe2019query}, song structure analysis \cite{watanabechorus} and other musical facets such as quality and release date \cite{smith2012your,o2012inferring,fell2014lyrics}. This makes them good candidate features for automatic music tagging (assigning labels like \textit{pop}, \textit{chill} to songs).



In the literature Hu and Downie \cite{hu2010lyrics} use collections of $n$-gram word counts (along with audio) for classifying mood.
Mayer et al. \cite{mayer2008rhyme} classify genre via rhyme analysis of lyrics, and Van Zaanen and Kanters \cite{van2010automatic} re-weight the word counts using TF-IDF (see \ref{sub:wordlevel}) to classify musical moods.
Text in these studies is often represented in \textit{Bag of Words} format \cite{hu2010lyrics,mayer2008rhyme,mayer2011musical}, where a vocabulary is built from a corpus and a song is represented as counts of the corpus words \cite{harris1954distributional}.
To obtain a usable vocabulary size, words are typically removed from the corpus if they appeared too often (stopwords such as \textit{the}, \textit{a}) or not often enough (bespoke vocabulary and misspellings).

Bag of Words is a useful intuitive document representation, but does not account for the fact that some words may have a low count in a document, yet still be considered interesting from a corpus perspective (for example, the word \textit{algorithm} in a corpus of agricultural documents).
Term Frequency Inverse Document Frequency (TF-IDF) \cite{jing2002improved} accounts for this by multiplying Bag of Words by a factor representing how common a word is in a corpus, and has also been explored in the music tagging context \cite{van2010automatic}.

The methods above have some clear drawbacks.
First, no semantic meaning is preserved or inferred between the individual words, meaning for example the model shares no information between words such as \textit{love} and \textit{adore}.
Second, the feature vectors can also easily become large and sparse (due to large vocabularies), making their use in machine learning models unwieldy.
Word2vec \cite{mikolov2013efficient} mitigates both of these issues by learning dense representations from a corpus, i.e. each word is represented as a point in a low-dimensional space in which semantically similar words are close.
The training objective in this model is to predict either a missing word given the context (Continuous Bag of Words) or vice-versa (Skipgram).
Word2vec has been adopted in a wide range of NLP tasks, including machine translation \cite{sutskever2014sequence}, sentiment analysis \cite{zhang2018deep}, and text generation \cite{brown2020language}; and in the Music Information Retrieval (MIR) domain has successfully been applied to explicit song detection \cite{rospocher2021explicit}, genre classification \cite{kumar2018genre} and music recommendation \cite{vystrvcilova2020lyrics}.

Although word order is considered in word2vec training, the algorithm does not provide a method for representing a document - something which is often needed for downstream tasks \cite{fell2014lyrics,smith2012your}.
One solution is to take summary statistics of the constituent word embeddings (i.e. simple averaging \cite{vystrvcilova2020lyrics}).
Another approach from the NLP literature is Doc2Vec \cite{le2014distributed}, which learns paragraph-level representations of documents via an additional model input representing paragraph indices.
Finally, it is possible to train the aggregation of word into document embeddings, for example using the final state of a recurrent neural network or the output of a self-attentive probe layer \cite{xun2019meshprobenet}.
Advanced models such as these were used by Alexandros Tsaptsinos \cite{tsaptsinos2017lyricsbased} to classify 20 music genres in a corpus of around 500,000 documents.

Solving these two problems (large vocabulary size and variable sequence lengths) is crucial to designing an accurate music tagging system from lyrics.
Making this work practically, and at scale, is the subject of this paper.
More concretely we investigate the efficaciousness off ``off-the-shelf'' language models trained on $\mathcal{O}(100B)$ tokens, training our own word embeddings from scratch on a bespoke lyrics dataset, and ``warm-starting'' the training.
To produce a representation of an entire song, we evaluate whether word-level features should be averaged, or processed using recurrent architectures.
It is out hope that this paper will serve as a practical guide for researchers hoping to make use of lyrics in tagging tasks.

\section{Methods}\label{sec:methods}
\begin{figure}
\centering
\includegraphics[width=\columnwidth]{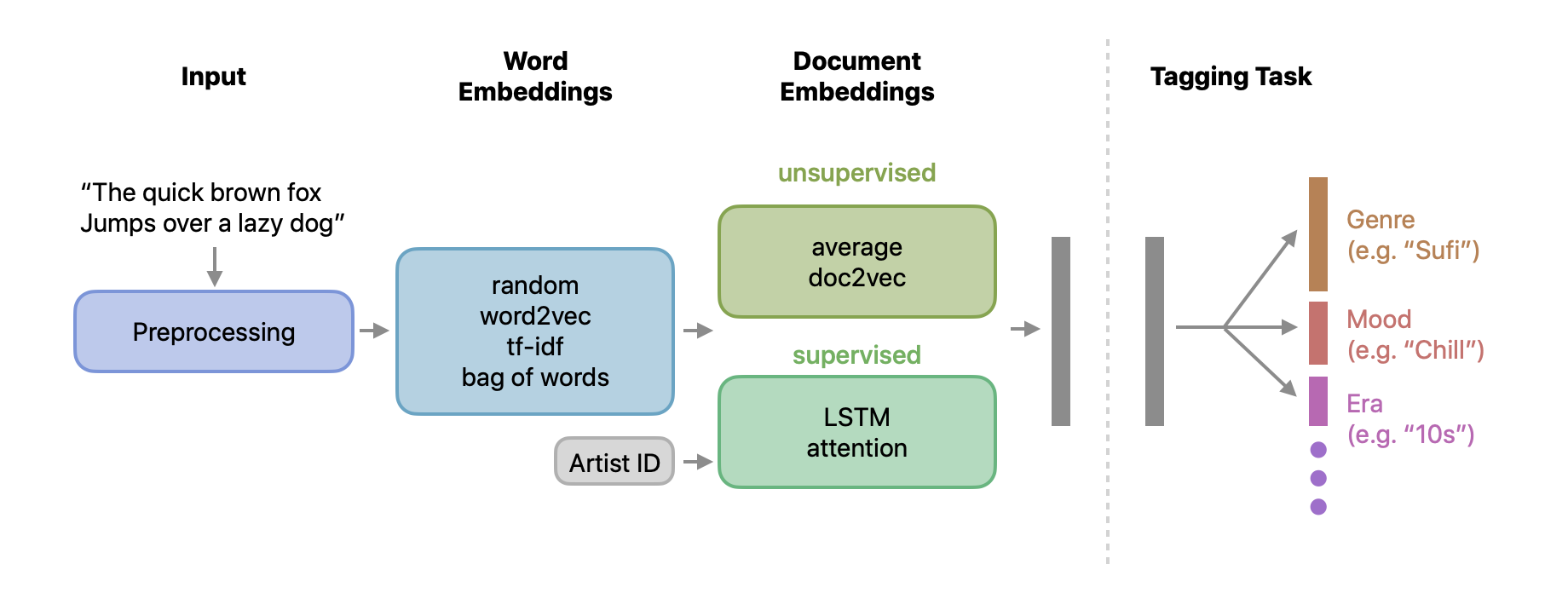}
\caption{We begin by processing raw lyric strings, before generating embeddings at the word level. We then have various methods of computing document embeddings: both supervised (sequence models with artist identification as the target) and unsupervised (averaging and doc2vec). At the end of this process we have a single embedding per document, which is our proposed representation. We then evaluate these embeddings by training deep tagging classifiers on the same representation.}
\label{fig:overview}
\end{figure}

The core of our investigation is trialling several options of representing song lyrics as an embedding. For this purpose we chose a transfer learning setup with distinct document embedding and tagging stages (Figure~\ref{fig:overview}). This setup has benefits beyond our investigation: the document representation can be learned from massive amounts of unlabelled lyrics, and can be re-used for different downstream tasks. We describe the model components in detail below, beginning with some definitions.
\subsection{Definitions}
In line with the NLP literature, we will refer to the \textit{lyrics to a song} as a \textit{document}, and to a \textit{collection of lyrics} as a \textit{corpus}.
A document is made up of multiple \textit{words}, usually broken by whitespace, but it is sometimes more convenient to work with subword \textit{tokens} so that information can be shared between words like \textit{play}, \textit{played}, and \textit{playing} in a model.
Broader structural information within documents come in the form of \textit{sentences} (groups of words separated by a period or other punctuation) or \textit{paragraphs} (longer groups of words separated by line breaks).
Lyrics do not often feature well-defined sentences but instead are arranged into \textit{lines} and \textit{stanzas}.
As these are roughly analogous to sentences/paragraphs, so we will refer to them as such in the remainder of the paper.

\subsection{Word embeddings}\label{sub:wordlevel}
\subsubsection{Baseline Models}
We begin by defining some simple baseline models:
\begin{itemize}
  \item \texttt{random}: random embeddings of dimension 128.
  \item \texttt{bag-of-words-d}: bag-of-words models with dimension $d$.
  \item \texttt{tf-idf-d}: TF-IDF models of dimension $d$.
\end{itemize}
For \texttt{bag-of-words-d}/\texttt{tf-idf-d}, we trimmed the vocabulary of the corpus by removing words which appeared in at least 90\% of documents, and then retained the $d$ most commonly occurring words. We had initially planned to reduce the dimensionality of the baseline models in a more principled way through dimensionality reduction techniques such as Principal Component Analysis, but realized that even with a sparse implementation we could not scale these techniques to our dataset size.

\subsubsection{Custom-trained word2vec}
Next, we trained word2vec models on our dataset, using the Python package \texttt{gensim}\footnote{\url{https://radimrehurek.com/gensim/}} to omit words which occurred fewer than five times in the dataset, and trained for 5 epochs -- these hyperparameters seemed sensible enough that we did not attempt to optimize them.
We did however try several embedding dimensions for use in downstream evaluation (see Table \ref{tab:hparams}), and refer to these models as \texttt{word2vec-d} for dimensionality $d$.

\subsubsection{Pre-trained word2vec}\label{subsub:pretrained}
The appeal of pretrained embeddings is that they have been exposed to a massive amount of text -- typically several orders of magnitude larger than in-house datasets.
They can therefore learn a good general-purpose understanding of word semantics, which can then optionally be fine-tuned on a specific domain task.
For our experiments we used the google news 300 dataset, which contains 300-dimensional vectors for around 3 million words, trained on around 100 Billion tokens \cite{mikolov2013efficient}.
Naturally some words appeared in our data for which no pretrained embedding existed - these were simply omitted.
We refer to this model as \texttt{google-300}.

\subsubsection{Warm-start word2vec}
We also attempted to ``warm-start'' the training of the embeddings from the model above into new embeddings \texttt{google-300-warm} - these vectors retained their dimensionality and we kept the same training hyper-parameters as \texttt{word2vec-d}.
The vocabularies of the two models were merged, such that words which appeared in both models took their initial state from \texttt{google-300} whilst words which were unique to \texttt{word2vec-d} had random initial state.

\subsection{Word Embedding Summaries}
\medskip For all representations above, to obtain a document-level representation we used averaging (for \texttt{word2vec-d}) or the native summary statistic (e.g. summing word counts in a document for \texttt{bag-of-words-d}).

In order to take paragraph structure and/or word order into account when computing document embeddings, we make use of more sophisticated summarization techniques.
This section investigates various methods for achieving this.

\subsubsection{doc2vec}
We begin with doc2vec \cite{le2014distributed}, once again using the \texttt{gensim} implementation. We refer to these models as \texttt{doc2vec-d}.

\subsubsection{LSTM and Attention}
Next, we kept the best-performing word embeddings from Subsection \ref{sub:wordlevel} and experimented with two neural sequence models: Long Short Term Memory networks (\texttt{lstm}), and an attention network (\texttt{attention}).
In order to learn the sequence parameters for these models, we needed a target for the model to predict.
Not wanting to use any labels which would be later used in our evaluation framework (see \ref{sub:tagging}), we decided to use the artist identifier as the target.

The number of unique artists in our dataset is naturally very large, so we considered using negative sampling \cite{mikolov2013distributed} to simplify the task for the networks.
However, we noticed in prior informal experiments that good results can actually be obtained with a large softmax layer instead.
Practically speaking, we proceeded by selecting the $1,000$ most common artists in the dataset and computing their song counts.
We then randomly sampled as many songs as we could for each of these artists such that we obtained a balanced dataset.
The final state for \texttt{lstm}, or the aggregated embedding for \texttt{attention}, were then connected to the target with dense layers.

We defer the discussion of results until Section \ref{sec:results}, but note here that both these models achieved a categorical accuracy in the artist identification proxy task of around 0.85\footnote{a random classifier would score around 0.001}.
In the next Subsection we describe our tagging model and the datasets used to evaluate and compare different document embeddings.

\subsection{Tagging Framework}\label{sub:tagging}
\subsubsection{Multi-label}
Our tagging model is a multi-task neural network architecture, where predictions on different tag vocabularies are treated as different tasks.
The input embedding is projected through a stack of fully connected layers until it branches to a number of linear output layers, one per tag vocabulary.
The loss function used to train the network is obtained by summing the binary cross-entropy loss terms associated with the output branches.
Note that binary cross-entropy loss is used instead of the categorical cross-entropy loss because multiple tags within the same vocabulary can be active for the same document.

\subsubsection{Multi-task}
We use multiple annotated datasets, defined over different set of documents: each dataset defines its own tag vocabulary and task.
The multi-task formulation makes it convenient to handle missing annotations, while still training all tasks in parallel.
The overall loss is:
\begin{equation}
\mathcal{L}_d = \sum_i \lambda_{i} a_{i,d} \mathcal{L}_{i,d},
\end{equation}
where $\mathcal{L}_{i,d}$ is the loss term associated with the $i$-th task for document $d$, $\lambda_i$ the loss weight for the task, and $a_{i,d} \in [0, 1]$ a binary flag that represents whether document $d$ is present in the annotations for task $i$.
When a track does not appear in an annotation dataset, the loss terms associated with that dataset is set to zero.

\subsubsection{Training}
During training, mean Average Precision (mAP) is computed at each epoch, and training is stopped when mAP reaches a plateau on the validation set.
Vocabulary-wise metrics are obtained simply by averaging the values for each tag, and a final scalar value is obtained by averaging across all tag vocabularies, weighted by the number of tags in each vocabulary. A summary of the hyper-parameters searched for all models is shown in Table \ref{tab:hparams}.



\section{Datasets}\label{sec:datasets}
\subsubsection{Lyrics datasets}\label{ssub:lyrics-datasets}

We began with an internal dataset of 17,389,303 documents with primary language as English.\footnote{deriving multiple language embeddings is an interesting extension of our work but beyond the scope of this paper}
Documents were then tokenized in \texttt{gensim} via the \texttt{simple\textunderscore preprocess} function.
We discovered that the distribution of number of tokens in the documents had an extremely long tail.
This was prohibitive for sequence models, so for all document embedding experiments we truncated the number of tokens to 512, which reduced the maximum sequence length from 8,641 to 512 yet only affected 4\% of documents.
After preprocessing, we were left with a corpus of approximately 3.8 billion tokens.

\subsubsection{Tagging datasets}\label{ssub:tagging-datasets}
We trained the tagging models on a set of internal datasets that were either manually curated or created from metadata.
The datasets contain tags from different domains, e.g. genre, mood, release date, and are defined over different, but overlapping, set of documents.
A description of the datasets is provided in Table \ref{tab:tagdatasets}.



\begin{table}[ht]
\centering
\setlength{\tabcolsep}{4pt}
\begin{tabular}{cccccccc}
\toprule
 & & & & & \multicolumn{3}{c}{Examples per label}\\
 \cmidrule(lr){6-8}
 Dataset & Example tag & Tracks & Tags & Tags/track & Min & Mean & Max\\
\midrule
Flagger & \texttt{spoken} & 58,444 & 6 & 1 & 2,700 & 9,740 & 39,796 \\
Era & \texttt{10s} & 50,450 & 9 & 1 & 87 & 5,769 & 17,616 \\
Moods & \texttt{Chill} & 71,271 & 22 & 1.9 & 100 & 6,092 & 21,995 \\
Explicit & \texttt{True} & 48,683 & 2 & 1 & 16,234 & 24,241 & 32,449 \\
Genre-1 & \texttt{Sufi} & 2,702,226 & 460 & 2.1 & 25 & 5,835 & 122,224 \\
Genre-2 & \texttt{Piano} & 479,792 & 273 & 1 & 490 & 1,757 & 2,000 \\
Genre-3 & \texttt{East Coast Rap} & 39,087 & 261 & 3.3 & 21 & 497 & 11,105 \\
Genre-4 & \texttt{Worship} & 562,274 & 25 & 3.6 & 152 & 79,966 & 426,008 \\
\bottomrule
\end{tabular}
\caption{\label{tab:tagdatasets}Tag datasets used for evaluation. Tags/track is simply the number of tags per track, averaged over each dataset.}
\end{table}

Some of the tag datasets may contain multiple labels for the same track, which makes creating balanced data splits more challenging. We used iterative splitting \cite{sechidis2011stratification}, while also forcing tracks from the same album to appear in the same split \cite{flexer09albumand}.
Note that some of the tagging datasets do overlap with the datasets used to train the document embeddings.
However the risk of overfitting here is small because the only label we use for training our embeddings is the artist identifier (see Subsection \ref{sub:wordlevel}).

\section{Results}\label{sec:results}
\begin{table}[ht]
\centering
\setlength{\tabcolsep}{4pt}
\begin{tabular}{lc}
\toprule
Hyperparameter & Values\\
\midrule
embedding dimension & \{$2^7, \ldots, 2^9$\}\\
dropout & \{$0.1, \ldots ,0.9$\}\\
learning rate & \{$10^{-1}, \ldots, 10^{-5}$\}\\
dense layers & \{$2^0, \ldots, 2^3$\}\\
dense size & \{$2^3, \ldots, 2^9$\}\\
lstm units & \{$2^6, \ldots, 2^9$\}\\
attention probes & \{$2^2, \ldots, 2^5$\}\\
attention mapping dimension & \{$2^2, \ldots, 2^5$\}\\
\bottomrule
\end{tabular}
\caption{\label{tab:hparams}Hyper-parameters evaluated in our experiments. Bayesian hyperparameter optimization \cite{snoek2012practical} was used to optimize the validation mean Average Precision, with early stopping and patience of 10 epochs. 20 trials were run concurrently and in total 100 trials were conducted for each model.
}

\end{table}
\subsection{Word embeddings}
\begin{figure}
\centering
\includegraphics{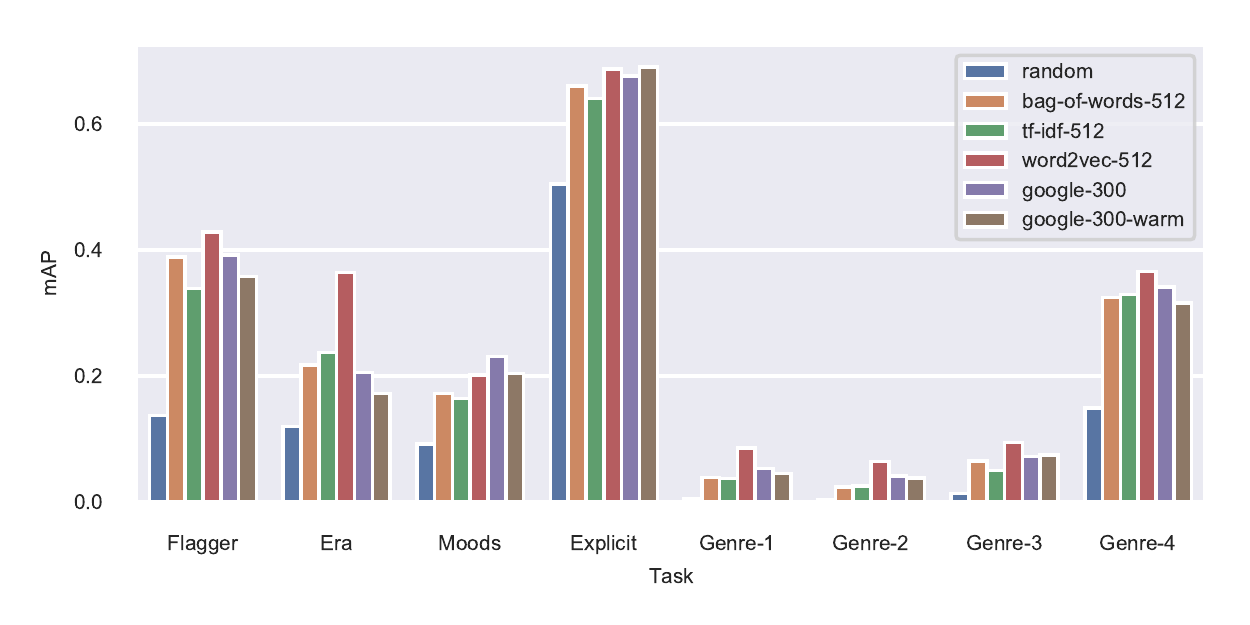}
\caption{\label{fig:word}Word-level embedding experiments, showing mAP on each tagging task. Document embeddings obtained by averaging/summing embeddings across words.}
\end{figure}

We show our results for overall mAP using word embeddings in Figure \ref{fig:word}, showing only the best-performing model dimensionality in each group.
All models outperform the random baseline but accuracy is varied across the tasks, owing in part to the differences in vocabulary size (recall Table \ref{tab:tagdatasets}).
The \texttt{word2vec-512} model with averaging achieves top performance on 6 of the 8 tasks, and is a close second on the \texttt{Flagger} task.

The only task on which a pretrained model is able to compete with \texttt{word2vec-512} is on the \texttt{Moods} dataset.
In general, warm-starting the training of embeddings did not yield improvements on our evaluation datasets.

\subsection{Document embeddings}
We selected \texttt{word2vec-512} as our best-performing word-level embedder, and set out to see if we could improve over simple embedding averaging -- see Table \ref{tab:doc} for our results.
Here we see that only \texttt{attention} is able to compete with \texttt{word2vec-512}, reaching similar performance on \texttt{Genre-3} and superior scores on the \texttt{Moods} and \texttt{Explicit} datasets.

Given the ability of \texttt{attention} to effectively label moods and explicit content, it seems that artist identification was a suitable proxy task for training the sequence models, or that the attention architecture is well suited for tasks related with specific keywords, such as emotions for moods or offensive content for \texttt{Explicit}.

It is unclear why the powerful \texttt{lstm/attention} models do not yield higher scores.
One reason could be that we have sufficient data to train excellent word embeddings, such that further refinements are simply hard to realize.
With this in mind, and knowing that in many cases large amounts of data are difficult to come by, we were interested to see what kind of performance could be attained from subsets of our data.
\begin{table}[ht]
\centering
\setlength{\tabcolsep}{4pt}
\begin{tabular}{ccccccccc}
\toprule
& Flagger & Era & Moods & Explicit & Genre-1 & Genre-2 & Genre-3 & Genre-4\\
\midrule
word2vec-512 & \textbf{0.429} & \textbf{0.365} & 0.202 & 0.687 & \textbf{0.086} & \textbf{0.065} & 0.095 & \textbf{0.366}\\
doc2vec-512 & 0.368 & 0.271 & 0.183 & 0.727 & 0.060 & 0.037 & 0.069 & 0.358\\
lstm & 0.330 & 0.247 & 0.204 & 0.723 & 0.044 & 0.041 & 0.057 & 0.282\\
attention & 0.427 & 0.295 & \textbf{0.272} & \textbf{0.760} & 0.070 & 0.057 & \textbf{0.107} & 0.350\\
\bottomrule
\end{tabular}
\caption{\label{tab:doc}Mean average precision for each model and tagging dataset for computing document embeddings. Best results for each dataset are in boldface.}
\end{table}

\subsection{Incremental training}
\begin{figure}
\centering
\includegraphics{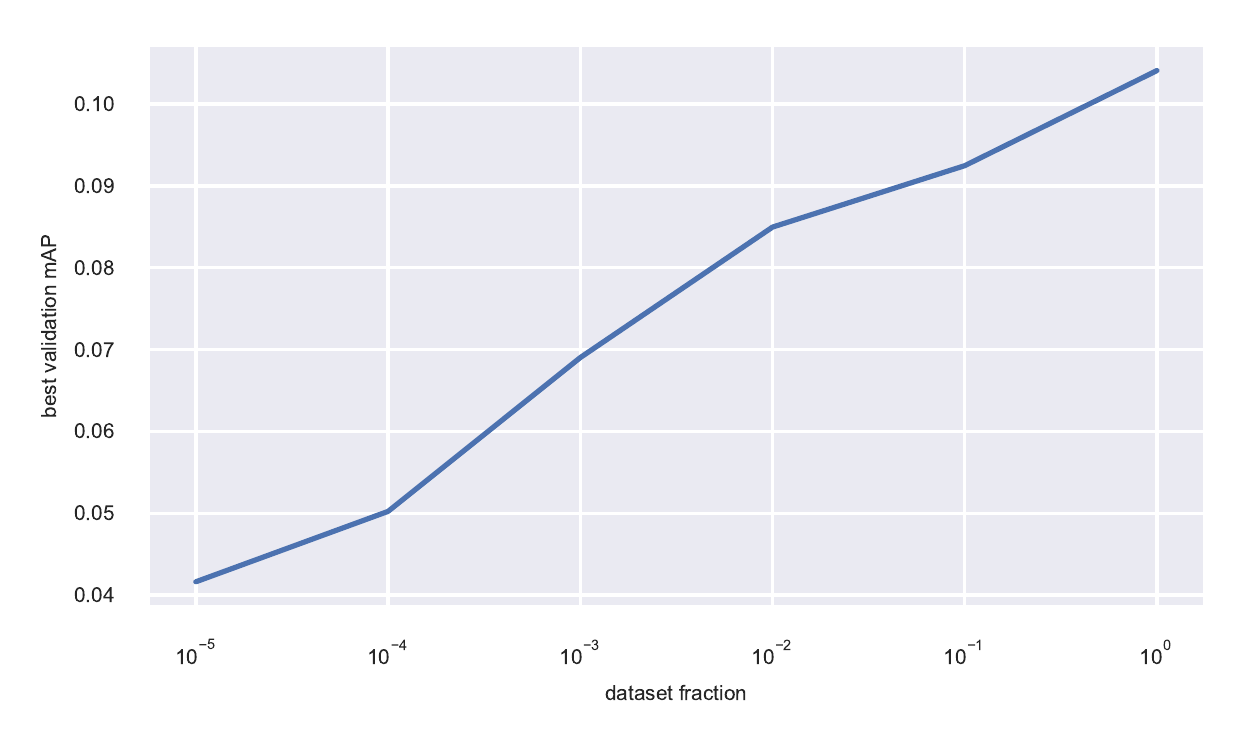}
\caption{Effect of the dataset sample size on overall mean Average Precision for the tagging task and \texttt{word2vec-512} model.}
\label{fig:incremental}
\end{figure}

We trained \texttt{word2vec-512} on random subsets of our data: 0.001\%, 0.01\%, 0.1\%, 1\%, 10\%, retaining the full evaluation test set in each case.
Results can be seen in Figure \ref{fig:incremental}, and show that in fact over 80\% of the mean average precision can be obtained from $1\%$ of the data (around 170,000 songs).

\section{Conclusions}\label{sec:conclusions}
In this paper, we provided a comprehensive quantitative analysis of word2vec style embeddings for music tagging.
On a range of challenging tagging tasks at the scale of millions of songs, we discovered that it is hard to surpass the performance of relatively simple models trained on in-house data.
Small improvements to averaging embeddings were shown to be possible through sequence modelling, although results were not conclusive.
Experiments on sampled data show that increasing training set size beyond $O$(1M) songs did not significantly improve tagging performance.

In future work, we are interested about the idea of extending our embedding framework to languages beyond English, and also seeing how useful our embeddings are as a source of side information in tasks such as music recommendation.


\bibliographystyle{plain}

\begin{thebibliography}{10}

\bibitem{brown2020language}
Tom~B Brown, Benjamin Mann, Nick Ryder, Melanie Subbiah, Jared Kaplan, Prafulla
  Dhariwal, Arvind Neelakantan, Pranav Shyam, Girish Sastry, Amanda Askell,
  et~al.
\newblock Language models are few-shot learners.
\newblock {\em arXiv preprint arXiv:2005.14165}, 2020.

\bibitem{delbouys2018music}
R{\'e}mi Delbouys, Romain Hennequin, Francesco Piccoli, Jimena Royo-Letelier,
  and Manuel Moussallam.
\newblock Music mood detection based on audio and lyrics with deep neural net.
\newblock {\em arXiv preprint arXiv:1809.07276}, 2018.

\bibitem{fell2014lyrics}
Michael Fell and Caroline Sporleder.
\newblock Lyrics-based analysis and classification of music.
\newblock In {\em Proceedings of COLING 2014, the 25th international conference
  on computational linguistics: Technical papers}, pages 620--631, 2014.

\bibitem{flexer09albumand}
Arthur Flexer and Dominik Schnitzer.
\newblock Album and artist effects for audio similarity at the scale of the
  web.
\newblock In {\em Proceedings of the 6th Sound and Music Computing Conference
  (SMC-09}, 2009.

\bibitem{harris1954distributional}
Zellig~S. Harris.
\newblock Distributional structure.
\newblock {\em WORD}, 10(2-3):146--162, 1954.

\bibitem{hu2010improving}
Xiao Hu and J~Stephen Downie.
\newblock Improving mood classification in music digital libraries by combining
  lyrics and audio.
\newblock In {\em Proceedings of the 10th annual joint conference on Digital
  libraries}, pages 159--168, 2010.

\bibitem{hu2010lyrics}
Xiao Hu and J~Stephen Downie.
\newblock When lyrics outperform audio for music mood classification: A feature
  analysis.
\newblock In {\em ISMIR}, pages 619--624, 2010.

\bibitem{hu2009lyric}
Xiao Hu, J~Stephen Downie, and Andreas~F Ehmann.
\newblock Lyric text mining in music mood classification.
\newblock {\em American music}, 183(5,049):2--209, 2009.

\bibitem{jing2002improved}
Li-Ping Jing, Hou-Kuan Huang, and Hong-Bo Shi.
\newblock Improved feature selection approach tfidf in text mining.
\newblock In {\em Proceedings. International Conference on Machine Learning and
  Cybernetics}, volume~2, pages 944--946. IEEE, 2002.

\bibitem{kumar2018genre}
Akshi Kumar, Arjun Rajpal, and Dushyant Rathore.
\newblock Genre classification using word embeddings and deep learning.
\newblock In {\em 2018 International Conference on Advances in Computing,
  Communications and Informatics (ICACCI)}, pages 2142--2146. IEEE, 2018.

\bibitem{laurier2008multimodal}
Cyril Laurier, Jens Grivolla, and Perfecto Herrera.
\newblock Multimodal music mood classification using audio and lyrics.
\newblock In {\em 2008 Seventh International Conference on Machine Learning and
  Applications}, pages 688--693. IEEE, 2008.

\bibitem{le2014distributed}
Quoc Le and Tomas Mikolov.
\newblock Distributed representations of sentences and documents.
\newblock In {\em International conference on machine learning}, pages
  1188--1196. PMLR, 2014.

\bibitem{mayer2008combination}
Rudolf Mayer, Robert Neumayer, and Andreas Rauber.
\newblock Combination of audio and lyrics features for genre classification in
  digital audio collections.
\newblock In {\em Proceedings of the 16th ACM international conference on
  Multimedia}, pages 159--168, 2008.

\bibitem{mayer2008rhyme}
Rudolf Mayer, Robert Neumayer, and Andreas Rauber.
\newblock Rhyme and style features for musical genre classification by song
  lyrics.
\newblock In {\em ISMIR}, pages 337--342, 2008.

\bibitem{mayer2011musical}
Rudolf Mayer and Andreas Rauber.
\newblock Musical genre classification by ensembles of audio and lyrics
  features.
\newblock In {\em ISMIR}, pages 675--680, 2011.

\bibitem{mckay2010evaluating}
Cory McKay, John~Ashley Burgoyne, Jason Hockman, Jordan~BL Smith, Gabriel
  Vigliensoni, and Ichiro Fujinaga.
\newblock Evaluating the genre classification performance of lyrical features
  relative to audio, symbolic and cultural features.
\newblock In {\em ISMIR}, pages 213--218, 2010.

\bibitem{mikolov2013efficient}
Tomas Mikolov, Kai Chen, Greg Corrado, and Jeffrey Dean.
\newblock Efficient estimation of word representations in vector space.
\newblock {\em arXiv preprint arXiv:1301.3781}, 2013.

\bibitem{mikolov2013distributed}
Tomas Mikolov, Ilya Sutskever, Kai Chen, Greg Corrado, and Jeffrey Dean.
\newblock Distributed representations of words and phrases and their
  compositionality.
\newblock {\em arXiv preprint arXiv:1310.4546}, 2013.

\bibitem{o2012inferring}
Tom O'Hara, Nico Sch{\"u}ler, Yijuan Lu, and Dan Tamir.
\newblock Inferring chord sequence meanings via lyrics: Process and evaluation.
\newblock In {\em ISMIR}, pages 463--468, 2012.

\bibitem{rospocher2021explicit}
Marco Rospocher.
\newblock Explicit song lyrics detection with subword-enriched word embeddings.
\newblock {\em Expert Systems with Applications}, 163:113749, 2021.

\bibitem{sechidis2011stratification}
Konstantinos Sechidis, Grigorios Tsoumakas, and Ioannis Vlahavas.
\newblock On the stratification of multi-label data.
\newblock In {\em Proceedings of the 2011 European Conference on Machine
  Learning and Knowledge Discovery in Databases}, ECML PKDD'11, page 145–158,
  2011.

\bibitem{smith2012your}
Alex~G Smith, Christopher~XS Zee, and Alexandra~L Uitdenbogerd.
\newblock In your eyes: Identifying clich{\'e}s in song lyrics.
\newblock In {\em Proceedings of the Australasian Language Technology
  Association Workshop 2012}, pages 88--96, 2012.

\bibitem{snoek2012practical}
Jasper Snoek, Hugo Larochelle, and Ryan~P Adams.
\newblock Practical bayesian optimization of machine learning algorithms.
\newblock {\em arXiv preprint arXiv:1206.2944}, 2012.

\bibitem{sutskever2014sequence}
Ilya Sutskever, Oriol Vinyals, and Quoc~V Le.
\newblock Sequence to sequence learning with neural networks.
\newblock {\em arXiv preprint arXiv:1409.3215}, 2014.

\bibitem{tsaptsinos2017lyricsbased}
Alexandros Tsaptsinos.
\newblock Lyrics-based music genre classification using a hierarchical
  attention network.
\newblock In {\em ISMIR}, 2017.

\bibitem{van2010automatic}
Menno Van~Zaanen and Pieter Kanters.
\newblock Automatic mood classification using tf* idf based on lyrics.
\newblock In {\em ISMIR}, pages 75--80, 2010.

\bibitem{vystrvcilova2020lyrics}
Michaela Vystr{\v{c}}ilov{\'a} and Ladislav Pe{\v{s}}ka.
\newblock Lyrics or audio for music recommendation?
\newblock In {\em Proceedings of the 10th International Conference on Web
  Intelligence, Mining and Semantics}, pages 190--194, 2020.

\bibitem{watanabechorus}
Kento Watanabe and Masataka Goto.
\newblock A chorus-section detection method for lyrics text.
\newblock pages 351--359.

\bibitem{watanabe2019query}
Kento Watanabe and Masataka Goto.
\newblock Query-by-blending: A music exploration system blending latent vector
  representations of lyric word, song audio, and artist.
\newblock In {\em ISMIR}, pages 144--151, 2019.

\bibitem{xun2019meshprobenet}
Guangxu Xun, Kishlay Jha, Ye~Yuan, Yaqing Wang, and Aidong Zhang.
\newblock Meshprobenet: a self-attentive probe net for mesh indexing.
\newblock {\em Bioinformatics}, 35(19):3794--3802, 2019.

\bibitem{yang2009music}
Dan Yang and Won-Sook Lee.
\newblock Music emotion identification from lyrics.
\newblock In {\em 2009 11th IEEE International Symposium on Multimedia}, pages
  624--629. IEEE, 2009.

\bibitem{ying2012genre}
Teh~Chao Ying, Shyamala Doraisamy, and Lili~Nurliyana Abdullah.
\newblock Genre and mood classification using lyric features.
\newblock In {\em 2012 International Conference on Information Retrieval \&
  Knowledge Management}, pages 260--263. IEEE, 2012.

\bibitem{zhang2018deep}
Lei Zhang, Shuai Wang, and Bing Liu.
\newblock Deep learning for sentiment analysis: A survey.
\newblock {\em Wiley Interdisciplinary Reviews: Data Mining and Knowledge
  Discovery}, 8(4):e1253, 2018.

\end{thebibliography}

\end{document}